\documentclass[10pt,twocolumn,letterpaper]{article}
\usepackage[pagenumbers]{cvpr}

\definecolor{cvprblue}{rgb}{0.21,0.49,0.74}
\usepackage[pagebackref,breaklinks,colorlinks,allcolors=cvprblue]{hyperref}
\usepackage{float}

\title{VMatcher: State-Space Semi-Dense Local Feature Matching}
\author{Ali Youssef}
\begin{document}
\maketitle
\begin{abstract}
This paper introduces VMatcher, a hybrid Mamba-Transformer network for semi-dense feature matching between image pairs. Learning-based feature matching methods, whether detector-based or detector-free, achieve state-of-the-art performance but depend heavily on the Transformer's attention mechanism, which, while effective, incurs high computational costs due to its quadratic complexity. In contrast, Mamba introduces a Selective State-Space Model (SSM) that achieves comparable or superior performance with linear complexity, offering significant efficiency gains. VMatcher leverages a hybrid approach, integrating Mamba's highly efficient long-sequence processing with the Transformer's attention mechanism. Multiple VMatcher configurations are proposed, including hierarchical architectures, demonstrating their effectiveness in setting new benchmarks efficiently while ensuring robustness and practicality for real-time applications where rapid inference is crucial. Source Code is available at: \url{https://github.com/ayoussf/VMatcher}
\end{abstract}    
\section{Introduction}
\label{sec:intro}
Finding reliable and accurate feature matches between image pairs is a core component in a multitude of Computer Vision tasks such as Structure-from-Motion (SfM), visual Simultaneous Localization and Mapping (vSLAM), Visual Localization, etc. Typically, feature detection and matching consist of three phases: feature detection, where distinctive interest points are extracted from an image; feature description, which involves encoding the local neighbourhood around the feature point into a stable high-dimensional descriptor vector; and feature matching, where descriptors from two images are paired to establish point-to-point correspondences, often through the mutual nearest-neighbour algorithm.

Traditionally, this problem has been approached with handcrafted methods, where feature detectors and descriptors such as \cite{SIFT, ORB, SURF} played a dominant role. These early methods relied on manually designed criteria to identify keypoints and establish matches, making them computationally efficient and interpretable. However, they often struggled under varying conditions, such as significant variation in viewpoint, illumination, and textureless regions, thus limiting their reliability in complex real-world scenarios. Over the past decade, learning-based methods have emerged, demonstrating greater robustness and consistently outperforming classical handcrafted approaches. This paradigm shift emphasises the importance of developing effective models and training strategies to maximise the potential of learning-based approaches.

In particular, the emergence of Transformers \cite{transformer} marked a breakthrough in feature detection and matching, leveraging their attention mechanism to capture complex global dependencies across images. This capability enabled both detector-based and detector-free matching methods to achieve remarkable accuracy and performance. However, the quadratic complexity of attention with respect to the sequence length imposes a significant computational burden, challenging real-time and resource-limited applications. Recently, Mamba \cite{Mamba} introduced a Selective State-Space Model (SSM) designed for efficient handling of long sequences with linear complexity, achieving performance on par with or surpassing Transformers in language tasks. Mamba's efficiency stems from a novel selection mechanism that dynamically adjusts its state-space parameters based on the input data, enabling effective modelling of complex patterns without the computational overhead associated with the attention mechanism. Nonetheless, Mamba’s autoregressive design limits its effectiveness in computer vision tasks, where spatial relationships are inherently parallel rather than sequential, and capturing global context is essential for accurate analysis \cite{MambaVision}.

Building on these insights, VMatcher is proposed, a hybrid Mamba-Transformer network for semi-dense feature matching that merges the strengths of Mamba's efficient linear processing capabilities with the Transformer's powerful attention mechanism. This hybrid approach effectively addresses computational bottlenecks while preserving the benefits of global context modelling, overcoming the ineffectiveness associated with Mamba's autoregressive formulation for spatial data, and mitigating the high computational costs inherent to Transformer architectures.

VMatcher introduces multiple design patterns and configurations, which achieve state-of-the-art accuracy on par with existing methods while improving computational efficiency, making it highly practical for real-world applications where rapid inference is essential. This work aims to strike a balance between robustness, speed, and accuracy in semi-dense matching, to establish VMatcher as a versatile and scalable solution for a diverse range of Computer Vision tasks.
\section{Related Work}\label{sec:related_work}

\subsection{Detector-Based Feature Matching} 

Traditional feature detection and matching methods, such as \cite{SIFT,SURF,ORB}, relied on handcrafted algorithms to identify repeatable, distinctive feature points and matched them using techniques such as nearest-neighbour (NN) search and mutual nearest neighbours (MNN). Despite their wide use, these classical methods often struggle with large variations in viewpoint, illumination, and appearance. To overcome these limitations, recent approaches have adopted learning-based methods, which deliver significantly better performance. SuperPoint \cite{SuperPoint} performs joint detection and description within a self-supervised Convolutional Neural Network (CNN) framework to improve robustness; however, it requires a complex training setup and high computational cost at large image scales. Following SuperPoint, other works \cite{D2-Net, R2D2, DISK, SiLK} have also leveraged CNN-based architectures for feature detection and description. Recently, XFeat \cite{XFeat} introduced a lightweight architecture that optimises both speed and accuracy for resource-efficient feature detection and matching, offering an ideal balance for real-time applications. 

Additionally, learning-based matchers such as SuperGlue \cite{SuperGlue} use Graph Neural Networks (GNNs) to ensure consistent correspondences by modelling relationships between feature points; however, it can be computationally expensive due to its quadratic complexity relative to the number of keypoints. In contrast, LightGlue \cite{LightGlue} adopts an adaptive matching scheme that balances efficiency and accuracy by adjusting computation based on matching difficulty, allowing for faster processing while maintaining competitive accuracy.

\subsection{Detector-Free Feature Matching}

Detector-free approaches perform direct feature matching without explicit keypoint detection, improving flexibility and robustness in challenging scenarios. LoFTR \cite{LoFTR} adopts a coarse-to-fine transformer framework, computing semi-dense correspondences to capture spatial relationships across an image pair without keypoint detection. While this approach results in high-quality matches, it suffers from significant computational overhead due to the need to compute the Transformer's \cite{transformer} attention over the entire feature map.

Subsequent works addressed efficiency limitations through various techniques. AspanFormer \cite{Aspanformer} introduced a hierarchical attention mechanism that dynamically adjusts the attention span based on flow maps, thereby reducing computational overhead while preserving matching accuracy. Similarly, MatchFormer \cite{Matchformer} offers a more refined matching process which follows an extract-and-match methodology on multiple feature scales. QuadTree Attention \cite{QuadTree} introduces a hierarchical attention mechanism that reduces computational cost without compromising performance. Rather than computing attention maps globally, it focuses on smaller, more relevant sections of the feature map, making it particularly effective for high-resolution matching tasks where managing computational load is crucial. Lastly, ELoFTR \cite{Eloftr} enhances the LoFTR framework \cite{LoFTR} through architectural refinements and the introduction of a novel transformer architecture that significantly improves efficiency while maintaining matching performance.

\subsection{Mamba} 

The Mamba architecture \cite{Mamba} introduces a Selective Scan Mechanism, designed to effectively balance local and global feature extraction while maintaining flexibility for various tasks. Initially developed for language modelling, Mamba has since been adapted and evolved into several variants for computer vision applications, including Vim \cite{vim}, which incorporated bidirectional encoding to improve spatial context, and EfficientVMamba \cite{EMamba}, which combined SSMs with CNN layers for a hierarchical scanning approach. VMamba \cite{VMamba} introduced the Cross-Scan Module to increase the receptive field, though it remains constrained by the directional scanning paths. 

More recently, MambaVision \cite{MambaVision} introduced several modifications to the Mamba block, addressing its limitations in computer vision tasks. In particular, it replaces causal convolutions with standard convolutions and introduces a symmetric branch with a convolution and SiLU activation function. With its parallel architecture for both sequential and spatial processing, and the ability to model both short- and long-range dependencies, MambaVision \cite{MambaVision} offers a rich feature representation with high efficiency, making it an ideal backbone for VMatcher's architecture.

\begin{figure*}
    \centering
    \includegraphics[width=\linewidth]{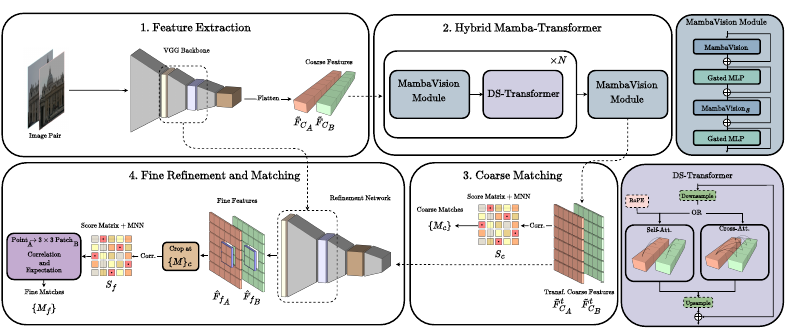}
    \caption{\textbf{VMatcher pipeline overview.} 1) VGG-style backbone extracts multi-scale feature maps from images $I_A$ and $I_B$, including coarse feature maps ${\tilde{F}_{C_A}}$ and ${\tilde{F}_{C_B}}$ at $\frac{1}{8}$ resolution. 2) Hybrid Mamba-Transformer module processes these for enhanced discrimination. 3) Coarse-level matches \{$M_c$\} obtained via correlation and MNN matching. 4) Fine-level feature maps are generated by fusing ${\tilde{F}_{C_A}}$ and ${\tilde{F}_{C_B}}$ with $\frac{1}{4}$ and $\frac{1}{2}$ resolution backbone features maps. Two-stage refinement: extracting patches centred on \{$M_c$\}, correlating via MNN, and aligning each point from $I_A$ with a $3 \times 3$ patch in $I_B$. Final-level matches \{$M_f$\} derived via softmax and expectation computation.}
    \label{fig:main-pipeline}
\end{figure*}
\section{Methodology}
\label{sec:methodology}

This section presents VMatcher’s methodology and model architecture, where given a pair of images $I_A$ and $I_B$, the objective is to establish accurate correspondences between an image pair efficiently. \cref{fig:main-pipeline} provides an overview of VMatcher's pipeline.

\subsection{Feature Extraction}\label{FeatureExtSec}

VMatcher adopts a lightweight and compact VGG-style feature extractor to maximise efficiency without compromising performance. The backbone consists of three sequential groups, each containing three blocks of Conv2D-BatchNorm-ReLU layers. Coarse feature maps ${\tilde{F}_{C_A}}$ and ${\tilde{F}_{C_B}}$ are extracted at $\frac{1}{8}$ resolution, along with additional finer feature maps at $\frac{1}{4}$ and $\frac{1}{2}$ resolutions utilised during the Fine Refinement and Matching stage. Lastly, convolutional and batch normalisation layers are fused during inference to minimise computational overhead. In contrast to previous work \cite{LoFTR,Eloftr,Aspanformer,Matchformer}, a lightweight extractor is favoured; for instance, despite having fewer parameters than comparable methods such as ELoFTR \cite{Eloftr} RepVGG \cite{RepVGG} feature extractor (1.5M vs. 9.5M), VMatcher's lightweight feature extractor maintains effectiveness while improving computational efficiency, as detailed in \cref{ablation}.

\subsection{Mamba}

\subsubsection{Preliminaries}
State-space models (SSMs) serve as a framework for representing continuous-time dynamics through two primary equations: the state transition equation and the observation equation. These equations define the behaviour of the system over time, and are based on a latent state $h(t)$ and state derivative $h'(t)$ that adjusts relative to input $x(t)$, with an output $y(t)$:

\begin{equation}
    h'(t) = \mathbf{A} h(t) + \mathbf{B} x(t).
\end{equation}

\begin{equation}
    y(t) = \mathbf{C} h(t).
\end{equation}

\noindent Where matrices $\mathbf{A}$, $\mathbf{B}$, and $\mathbf{C}$ determine the system dynamics in continuous time.

\noindent \textbf{Discretisation.} To apply continuous models to discrete data, the system equations must be discretised. Mamba \cite{Mamba} leverages the Zero-Order Hold (ZOH) method, using a sampling interval $\Delta$ to map the continuous matrices $\mathbf{A}$ and $\mathbf{B}$ into their discrete counterparts, $\bar{\mathbf{A}}$ and $\bar{\mathbf{B}}$:

\begin{equation} \bar{\mathbf{A}} = \exp(\Delta \mathbf{A}). \end{equation}

\begin{equation} \bar{\mathbf{B}} = (\Delta \mathbf{A})^{-1} \left( \exp(\Delta \mathbf{A}) - \mathbf{I} \right) \cdot \Delta \mathbf{B}. \end{equation}

This transformation allows the model to process sequential or other discrete inputs efficiently, aligning with modern digital computation \cite{ConvCombine}.

\noindent \textbf{Convolutional Computation.} Mamba \cite{Mamba} accelerates sequence processing by reformulating the discretised system into a convolutional operation using discretised matrices $\bar{\mathbf{A}}$ and $\bar{\mathbf{B}}$ rather than sequentially computing outputs, thus enhancing parallelisation. The output $y_i$ at time step $i$ is computed as:

\begin{equation}
\begin{aligned}
    y_i = \sum_{i=0}^{L} \mathbf{C} \bar{\mathbf{A}}^{L-i} \bar{\mathbf{B}} x_i.
\end{aligned}
\end{equation}

By constructing a set of convolutional kernels $\mathbf{\bar K} = \left( \mathbf{C} \bar{\mathbf{B}}, \mathbf{C} \bar{\mathbf{A}} \bar{\mathbf{B}}, \dots, \mathbf{C} \bar{\mathbf{A}}^{L-1} \bar{\mathbf{B}} \right)$, the output is efficiently computed as:

\begin{equation}
y = x \ast \mathbf{\bar{K}}.
\end{equation}

\noindent Where $x = [x_0, x_1, \dots]$ and $y = [y_0, y_1, \dots] \in \mathbb{R}^L$, with $L$ denoting the sequence length. This convolutional structure improves efficiency by enabling parallel matrix operations, improving performance on high-dimensional data.

\noindent \textbf{Selection Mechanism.} Traditional SSMs lack the flexibility to adjust outputs based on inputs due to inherent time invariance. Mamba \cite{Mamba} overcomes this by introducing an input-dependent selection mechanism that generates dynamic time-varying weight matrices, allowing it to filter irrelevant information while preserving details relevant to each input.

\subsubsection{MambaVision}\label{VisionMambasec}

\begin{figure}
  \centering
  \begin{subfigure}{0.45\columnwidth}
    \includegraphics[width=\textwidth]{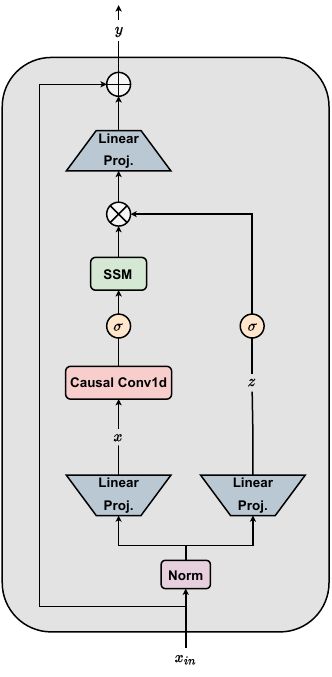}
    \caption{Vanilla Mamba \cite{Mamba} Block}
    \label{fig:short-a}
  \end{subfigure}
  \hfill
  \begin{subfigure}{0.45\columnwidth}
    \includegraphics[width=\textwidth]{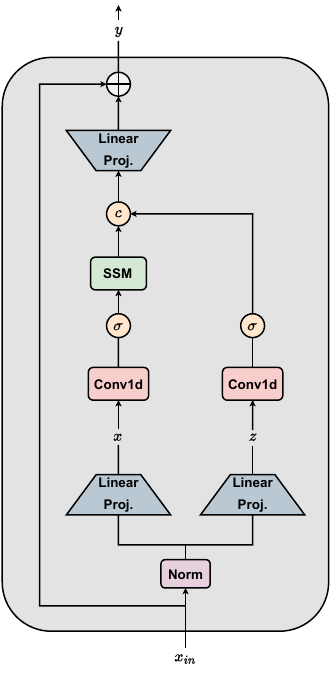}
    \caption{MambaVision \cite{MambaVision} Block }
    \label{fig:short-b}
  \end{subfigure}
  \caption{Mamba Block Comparison: (a) Vanilla Mamba \cite{Mamba} Block vs. (b) MambaVision \cite{MambaVision} Block. The MambaVision Block enhances the Vanilla Mamba by integrating a 1D convolution along with an additional symmetric branch, addressing the limitations of causal convolutions, thus improving performance in vision tasks.}
  \label{fig:VisionMambaBlock}
\end{figure}

MambaVision \cite{MambaVision} modifies vanilla Mamba \cite{Mamba} to enhance feature representation and generalisation in vision tasks. As shown in \cref{fig:VisionMambaBlock}, the causal 1D convolution is replaced with a 1D convolution to eliminate directional limitations. A symmetric branch bypassing the selective scan mechanism is added, consisting of a 1D convolution layer with SiLU activation to recover information potentially lost due to sequential constraints imposed by the SSM. To maintain a parameter count similar to that of Vanilla Mamba \cite{Mamba}, inputs are projected to half the embedding dimension, and outputs are concatenated and projected via a final linear layer. The output $y$ is computed as:

\begin{equation}
x = \text{SSM}(\sigma(\text{Conv}(\text{Linear}(C, \frac{C}{2})(x_{\text{in}})))),
\end{equation}

\begin{equation}
z = \sigma(\text{Conv}(\text{Linear}(C, \frac{C}{2})(x_{\text{in}}))),
\end{equation}

\begin{equation}
y = \text{Linear}(C, C)(\text{Concat}(x, z)).
\end{equation}

% This design enhances feature representation, generalisation, and performance on computer vision tasks.

\subsection{Downsampled-Transformer}\label{dstransformersec}

Following the MambaVision module, feature vectors $\tilde{F}_{C_A}$ and $\tilde{F}_{C_B}$ are passed through the Downsampling Attention Module that computes self-attention or cross-attention. Standard attention mechanism combines queries $Q$, keys $K$, and values $V$ through weighted sums:

\begin{equation}
  Attention(Q, K, V) = softmax(\frac{QK^T}{\sqrt{d_k}})V.
  \label{eq:attention}
\end{equation}

\noindent Where $d_k$ is the dimensionality of the feature vector, scaling the dot products for stability.

This form of attention effectively models dependencies between elements but can be computationally intensive. To mitigate computational costs, previous work has explored utilising linear attention \cite{LoFTR}, token-pyramid downsampling transformers \cite{QuadTree}, or ELoFTR's \cite{Eloftr} aggregated attention module that uses depth-wise convolution on the query-generating vector and max-pooling on the key- and value-generating vectors. However, VMatcher leverages Mamba's \cite{Mamba} selection scan mechanism to filter irrelevant information, eliminating the need for intricate techniques employed by prior methods. As such, by simply downsampling ${\tilde{F}_{C_A}}$ and ${\tilde{F}_{C_B}}$ using bilinear interpolation before attention computation and upsampling afterwards, this approach minimises model parameters while maintaining performance. 

\begin{figure}[h]
    \centering
    \includegraphics[width=0.5\columnwidth]{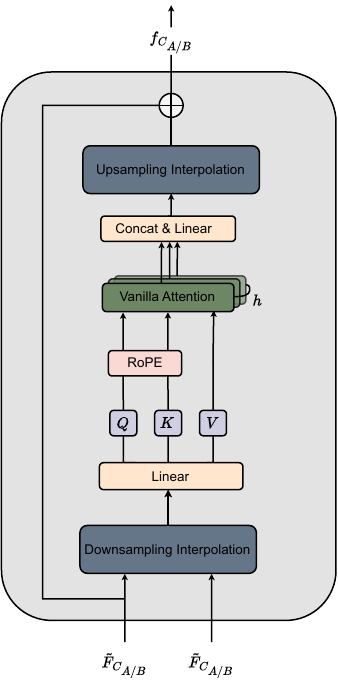}
    \caption{\textbf{Downsampled Transformer illustration.}}
    \label{fig:DStransformer}
\end{figure}

Moreover, the DS-Transformer module omits the typical Multi-Layer Perceptron (MLP) post-attention layer in favour of directly adding the attention vector to the input, as it led to degraded results (\emph{see} \cref{ablation}). Lastly, similar to previous work \cite{LightGlue,LoFTR,Eloftr,Aspanformer,Matchformer}, Rotary Positional Embedding (RoPE) \cite{Rope} is employed for queries and keys during self-attention to capture spatial dependencies. However, as Mamba's \cite{Mamba} sequence modelling properties and parameter structure \cite{Jamba, MambaVision} inherently encode positional information, utilising RoPE \cite{Rope} may not be necessary, which is further detailed in \cref{ablation}. An illustration of the DS-Transformer can be seen in \cref{fig:DStransformer}.

\subsection{Coarse-Level Matching}

Coarse-level correspondences are derived by densely correlating transformed feature vectors $\tilde{F}^t_{C_A}$ and $\tilde{F}^t_{C_B}$ to compute score matrix $S_C$, followed by a dual Softmax operation across both dimensions of $S_C$ to generate a matching probability map, as in \cite{LoFTR, Eloftr, DISK, SiLK}. Coarse-level matches \{$M_c$\} are obtained using mutual nearest neighbour (MNN) and a threshold $\tau$.

\subsection{Fine-Level Extraction and Refinement}

Adapted from ELoFTR \cite{Eloftr}, the fine-level refinement module efficiently achieves sub-pixel precision through two components:

\noindent\textbf{Fine Feature Extraction.} The transformed coarse feature vectors $\tilde{F}^t_{C_A}$ and $\tilde{F}^t_{C_B}$ are fused with backbone feature maps extracted at $\frac{1}{4}$ and $\frac{1}{2}$ resolutions through convolution and upsampling operations, yielding fine maps $\hat{F}_{f_A}$ and $\hat{F}_{f_B}$ at fine-level resolution. Fine-level feature patches centred on coarse matches \{$M_c$\} are then extracted from $\hat{F}_{f_A}$ and $\hat{F}_{f_B}$, allowing efficient yet precise refinement.

\noindent\textbf{Two-Stage Refinement.} First, dense correlation between fine-level feature patches produces a local patch score matrix $S_f$, where high-confidence pixel-level matches are identified via mutual nearest neighbour (MNN) matching. Second, each point in $I_A$ is correlated with a $3 \times 3$ feature patch centred at its matched location in $I_B$. Softmax is applied to the resulting match distribution matrix, and the refined matches \{$M_f$\} are obtained by computing the expectation.

\subsection{Supervision}

Following \cite{SuperGlue, LoFTR, Eloftr}, coarse-level supervision is achieved using ground truth matches \{$M_{c_{gt}}$\} generated via warping grid points from $I_A$ to $I_B$ using depth maps and camera poses. Coarse-level score matrix $S_C$ is supervised by minimising log-likelihood loss at \{$M_{c_{gt}}$\}:
\begin{equation}
  L_c = -\frac{1}{N} \sum_{(\tilde{i}, \tilde{j}) \in \{M_c\}_{gt}} \log S_c(\tilde{i}, \tilde{j}).
  \label{eq:coarse_supervision}
\end{equation}

Fine-level supervision adopts the two-stage approach of \cite{Eloftr}. $L_{f_{1}}$ minimises log-likelihood of each fine local score matrix $S_f$ using pixel-level ground truth matches, while $L_{f_{2}}$ computes $\ell_2$ loss between final sub-pixel \{$M_{f}$\} and ground truth \{$M_{f_{gt}}$\} matches as in \cite{Aspanformer, LoFTR, Eloftr}. The total loss is expressed as:
\begin{equation}
  L = L_c + \alpha L_{f_{1}} + \beta L_{f_{2}}.
  \label{eq:total_loss}
\end{equation}
\section{Implementation Details}
\label{sec:implementation_details}

\textbf{VMatcher Variations.} VMatcher offers two configurations: VMatcher-B (Base) with 9.5M parameters and VMatcher-T (Tiny) with 6.9M parameters; architecture details are provided in the Supplementary Material. Moreover, \cref{sec:experiments} includes \textbf{optimised} variants of both models, in which the Dual-Softmax operation is omitted during coarse-level matching, following \cite{Eloftr}. This modification greatly enhances inference time while maintaining performance.

\noindent \textbf{Inference Time Measurement.} Measuring inference time can be complex when comparing sparse and semi-dense methods. A direct comparison is slightly biased towards semi-dense methods since applications such as SfM extract feature points once and cache them, making inference time primarily a factor of feature matching. This does not extend to semi-dense methods, as caching large feature vectors is computationally intensive. Thus, for fairness, feature detection and matching times are separately reported for sparse methods in \cref{sec:experiments}.
\section{Experiments}\label{sec:experiments}

To ensure a fair evaluation, the LightGlue \cite{LightGlue} approach is adopted, tuning RANSAC's inlier threshold for each model on each test dataset, as the emphasis is on evaluating the models rather than RANSAC itself. \cref{tab:hpatches,tab:megadepth,tab:scannet} present results using PoseLib \cite{PoseLib} LO-RANSAC and OpenCV RANSAC \cite{ransac}, with runtimes marked \textbf{\emph{mp}} if mixed precision was used during inference. Dense methods are excluded despite strong performance, as their computational demands and runtime significantly exceed those of sparse and semi-dense approaches. For consistency, Flash-Attention \cite{FlashAtt} is omitted to ensure fair comparisons.

\subsection{Homography Estimation}\label{hom-est-sec}

Homography estimation is assessed on the HPatches dataset \cite{Hpatches}, which includes 57 and 59 planar scenes with varying illumination and viewpoints, each scene comprises 6 images linked by ground truth homography matrices.

\noindent\textbf{Baselines.} Comparisons are made against sparse methods, including SuperPoint \cite{SuperPoint} with Nearest Neighbours and SuperGlue \cite{SuperGlue}, as well as SuperPoint, DISK \cite{DISK}, and Aliked \cite{ALIKED} with LightGlue \cite{LightGlue}. For semi-dense methods, VMatcher is evaluated against QuadTree, Aspanformer, LoFTR, and ELoFTR \cite{QuadTree, LoFTR, Eloftr, Aspanformer}.

\noindent\textbf{Evaluation Protocol.} In line with prior work \cite{LoFTR, SuperGlue, SiLK}, the shorter side of each image is resized to 480 pixels. The mean projection error for the image's corner points is computed, and the area under the curve (AUC) error is reported at 3, 5, and 10 pixel thresholds. For consistency with sparse methods, results are reported using the top 1024 predicted matches.

\noindent\textbf{Results.} As shown in \cref{tab:hpatches}, VMatcher models outperform both sparse and semi-dense methods despite limiting keypoints to 1024. VMatcher-B matches ELoFTR \cite{Eloftr} in runtime (further analysed in \cref{Runtimebreakdown}), while VMatcher-T processes images 1.8$\times$ faster than LoFTR \cite{LoFTR} and 2.2$\times$ faster than AspanFormer \cite{Aspanformer} while maintaining competitive accuracy. The optimised variants further enhance efficiency, with VMatcher-T$_{\text{Opt}}$ reducing runtime by 16.9\%, achieving an optimal balance between accuracy and computational efficiency.

\begin{table}[ht]
\centering
\resizebox{\columnwidth}{!}{%
\begin{tabular}{ccccc}
\hline
Matcher Type         & Method       & \multicolumn{2}{c}{HPatches Dataset}          & Time (ms)                  \\ \cline{3-4}
                     &              & \multicolumn{2}{c}{AUC@3px / AUC@5px / AUC@10px}          &                            \\ \cline{3-4}
                     & & LO-RANSAC & RANSAC & \\ \hline
                     & SP+NN & 63.7 / 74.5 / 84.4 & 54.6 / 67.8 / 80.2 & \textbf{17.50} (15.58/2.36) \\
                     & SP+SG & \textbf{68.3} / \textbf{78.6} / \textbf{87.8} & 59.8 / 71.0 / 82.8 & 46.05 (15.58/30.47$_{mp}$) \\
Sparse               & SP+LG & 67.7 / 78.0 / 87.2 & 59.1 / 72.1 / 83.7 & 34.11 (15.14/18.97$_{mp}$) \\
                     & DISK+LG & 61.7 / 74.0 / 84.9 & 52.6 / 67.2 / 80.7 & 51.28 (33.42/17.85$_{mp}$) \\
                     & Aliked+LG & 66.3 / 77.1 / 86.7 & \textbf{61.1} / \textbf{73.4} / \textbf{84.5} & 36.74 (21.01/15.73$_{mp}$) \\ \hline
                     
                     & QuadTree & 69.5 / 79.2 / 87.3 & 67.4 / 77.4  / 85.9 & 100.64$_{mp}$              \\
\multicolumn{1}{l}{} & AspanFormer & 68.7 / 78.5 / 86.9 & 66.7 / 76.6 / 85.7 & 65.54$_{mp}$ \\
                     & LoFTR & 69.5 / 78.7 / 86.7 & 67.6 / 76.9 / 85.4 & 52.35$_{mp}$               \\
Semi-Dense           & ELoFTR & 69.6 / 78.9 / 87.1  & 67.5 / 77.3 / 86.0 & 34.48$_{mp}$ \\
                     & VMatcher-B & 69.8 / 79.1 / \textbf{87.5} & \textbf{67.8} / \textbf{77.5} / \textbf{86.4} & 35.43$_{mp}$ \\
                     & VMatcher-T & \textbf{70.2} / \textbf{79.3} / 87.4 & \textbf{67.8} / 77.3 / 86.0 & \textbf{29.23}$_{mp}$ \\ \hline
                     
                     & ELoFTR$_{\text{Opt}}$ & 68.6 / 78.2 / 86.7 & 65.6 / 75.4 / 84.5 & 28.62$_{mp}$ \\
Semi-Dense-Opt       & VMatcher-B$_{\text{Opt}}$ & \textbf{68.9} / \textbf{78.5} /\textbf{ 86.9} & \textbf{66.4} / \textbf{76.1} / \textbf{85.3} & 29.58$_{mp}$\\
                     & VMatcher-T$_{\text{Opt}}$ & 67.6 / 77.5 / 86.5 & 64.7 / 74.8 / 84.7 & \textbf{24.29}$_{mp}$\\ \hline
\end{tabular}%
}
\caption{\textbf{Homography Estimation Results on HPatches.} Reprojection error AUC scores and runtimes for sparse and semi-dense methods.}
\label{tab:hpatches}
\end{table}

\subsection{Relative Pose Estimation}\label{pose-est-sec}

Relative pose estimation evaluation is performed on both outdoor and indoor datasets. Following prior work \cite{LightGlue,SuperGlue,Eloftr,LoFTR} for outdoor scenes, the MegaDepth dataset \cite{Megadepth} test split from \cite{LoFTR} is utilised, which includes 1500 image pairs from "Sacre Coeur" and "St. Peter's Square". For indoor scenes, the ScanNet dataset \cite{Scannet} is used, which comprises 1613 sequences capturing challenges such as viewpoint variation and textureless regions. Evaluation is performed on the ScanNet \cite{Scannet} test split of 1500 image pairs from \cite{SuperGlue}.

\noindent\textbf{Baselines.} Relative pose estimation evaluation utilises similar baselines as in \cref{hom-est-sec}.

\noindent\textbf{Evaluation Protocol.} For outdoor scenes, images are resized with the longest edge at 1184 pixels for semi-dense methods and 1600 for sparse methods, as specified in \cite{LightGlue}. For indoor scenes, images are resized to 640x480 pixels. The pose error is measured as the maximum between angular rotation and translation errors, with results reported at thresholds of $5^\circ$, $10^\circ$, and $20^\circ$. Sparse methods use the top 2048 keypoints for evaluation. 

\noindent\textbf{Results.} As shown in \cref{tab:megadepth}, VMatcher variants outperform all sparse and semi-dense methods on MegaDepth \cite{Megadepth} while offering considerable runtime advantages. However, the performance differences between models are minimal ($\approx$1.2\% in AUC@$10^\circ$), with the primary distinction lying in computational efficiency, where VMatcher-B is 1.15$\times$ faster than ELoFTR~\cite{Eloftr} and 1.45$\times$ faster than LoFTR \cite{LoFTR}, while VMatcher-T achieves even greater speed improvements at 1.26 $\times$ and 1.60$\times$ faster, respectively. The optimised variants achieve near-sparse processing speeds with an insignificant degradation in performance.

From \cref{tab:scannet}, VMatcher performance is comparable to other semi-dense methods, with AspanFormer \cite{Aspanformer} leading in performance. Notably, VMatcher-B runtime is similar to ELoFTR \cite{LoFTR}, similar to findings in \cref{hom-est-sec} (further analysed in \cref{Runtimebreakdown}), while VMatcher-T offers additional runtime advantages. Interestingly, VMatcher-T$_{\text{Opt}}$ strikes the best balance between accuracy and runtime, outperforming all sparse methods in both aspects.

A key limitation to highlight is the instability in relative pose estimation, particularly in translation error computation. The error is derived from the angular difference between ground truth and estimated translation vectors, which becomes unreliable due to scale ambiguity \cite{Multi-View-G-CV, learning-good-corr}, especially with small ground-truth translation norms ($||t_{GT}||$). This affects 139 of 1500 image pairs in the ScanNet \cite{Scannet} test split.

\begin{table}[ht]
\centering
\resizebox{\columnwidth}{!}{%
\begin{tabular}{ccccc}
\hline
Matcher Type   & Method                    & \multicolumn{2}{c}{MegaDepth Dataset}                                                         & Time (ms)                                         \\ \cline{3-4}
               &                           & \multicolumn{2}{c}{AUC@$5^\circ$ / AUC@$10^\circ$ / AUC@$20^\circ$}                           &                                                   \\ \cline{3-4}
               &                           & LO-RANSAC                                     & RANSAC                                        &                                                   \\ \hline
               & SP+NN                     & 51.1 / 64.4 / 73.9                            & 39.1 / 54.3 / 66.4                            & \multicolumn{1}{c}{\textbf{43.58} (40.73/2.85)}  \\
               & SP+SG                     & 66.0 / 78.8 / 87.6                            & 50.7 / 68.2 / 80.9                            & \multicolumn{1}{c}{82.61 (40.73/41.88$_{mp}$)}   \\
Sparse         & SP+LG                     & \textbf{66.8} / \textbf{79.4} / \textbf{88.0} & 51.5 / \textbf{68.4} / \textbf{81.1}          & \multicolumn{1}{c}{58.25 (40.73/17.54$_{mp}$)}   \\
               & DISK+LG                   & 61.1 / 74.2 / 83.9                            & 44.7 / 62.5 / 76.4                            & \multicolumn{1}{c}{161.04 (143.67/17.36$_{mp}$)} \\
               & Aliked+LG                 & 66.3 / 78.9 / 87.5                            & \textbf{51.7} / 68.3 / 80.1                   & \multicolumn{1}{c}{80.04 (62.58/17.45$_{mp}$)}   \\ \hline
               & QuadTree                  & 66.5 / 78.9 / 87.2                            & 51.5 / 68.4 / 80.6                            & \multicolumn{1}{c}{398.38}                       \\
               & AspanFormer               & 69.2 / 80.9 / 88.7                            & 55.2 / 71.5 / 83.2                            & \multicolumn{1}{c}{160.73$_{mp}$}                \\
               & LoFTR                     & 68.3 / 80.1 / 88.3                            & 53.5 / 69.8 / 81.9                            & \multicolumn{1}{c}{126.94$_{mp}$}                \\
Semi-Dense     & ELoFTR                    & 69.4 / 80.8 / 88.7                            & \textbf{56.0} / 72.1 / \textbf{83.6}          & \multicolumn{1}{c}{100.25$_{mp}$}                \\
               & VMatcher-B                & \textbf{69.6} / \textbf{81.1} / \textbf{88.9} & \textbf{56.0} / \textbf{72.2} / \textbf{83.6} & \multicolumn{1}{c}{87.56$_{mp}$}                 \\
               & VMatcher-T                & 69.4 / 81.0 / 88.8                            & 55.8 / 71.9 / 83.4                            & \multicolumn{1}{c}{\textbf{79.46}$_{mp}$}        \\ \hline
               & ELoFTR$_{\text{Opt}}$     & 68.8 / 80.3 / 88.2                            & 55.9 / 71.9 / 83.3                            & \multicolumn{1}{c}{73.52$_{mp}$}                 \\
Semi-Dense-Opt & VMatcher-B$_{\text{Opt}}$ & \textbf{69.5} / \textbf{81.0} / \textbf{88.7} & \textbf{56.3} / \textbf{72.3} / \textbf{83.6} & \multicolumn{1}{c}{66.16$_{mp}$}                 \\
               & VMatcher-T$_{\text{Opt}}$ & 69.0 / 80.7 / 88.4                            & 55.5 / 71.6 / 83.2                            & \multicolumn{1}{c}{\textbf{55.94}$_{mp}$}        \\ \hline
\end{tabular}%
}
\caption{\textbf{Outdoor Relative Pose Estimation Results on the MegaDepth dataset \cite{Megadepth}.} VMatcher variants deliver improved performance and runtime speed advantages compared to semi-dense methods.}
\label{tab:megadepth}
\end{table}

\begin{table}[ht]
\centering
\resizebox{\columnwidth}{!}{%
\begin{tabular}{ccccc}
\hline
Matcher Type   & Method                    & \multicolumn{2}{c}{ScanNet Dataset}                                                           & Time (ms)                   \\ \cline{3-4}
               &                           & \multicolumn{2}{c}{AUC@$5^\circ$ / AUC@$10^\circ$ / AUC@$20^\circ$}                           &                             \\ \cline{3-4}
               &                           & LO-RANSAC                                     & RANSAC                                        &                             \\ \hline
               & SP+NN                     & 15.7 / 31.7 / 48.6                            & 11.5 / 25.6 / 42.3                            & \textbf{19.04} (14.56/4.48) \\
               & SP+SG                     & \textbf{22.1} / \textbf{40.8} / \textbf{58.1} & 18.1 / \textbf{36.6} / \textbf{54.4}          & 57.93 (14.56/43.47$_{mp}$)  \\
Sparse         & SP+LG                     & 20.0 / 38.0 / 55.6                            & 16.9 / 34.3 / 51.9                            & 42.79 (14.56/28.23$_{mp}$)  \\
               & DISK+LG                   & 18.8 / 34.1 / 49.2                            & \textbf{19.0} / 35.1 / 50.1                   & 48.16 (28.81/19.35$_{mp}$)  \\
               & Aliked+LG                 & 21.0 / 39.1 / 56.0                            & 17.0 / 33.6 / 50.3                            & 45.78 (23.76/22.02$_{mp}$)  \\ \hline
               & QuadTree                  & 24.8 / 43.8 / 60.3                            & 20.2 / 38.4 / 55.4                            & 89.77                       \\
               & AspanFormer               & \textbf{26.4} / \textbf{45.7} / \textbf{62.4} & \textbf{21.6} / \textbf{40.2} / \textbf{57.1} & 49.70$_{mp}$                \\
               & LoFTR                     & 22.4 / 39.8 / 55.6                            & 17.2 / 34.0 / 50.3                            & 33.02$_{mp}$                \\
Semi-Dense     & ELoFTR                    & 25.0 / 43.8 / 60.0                            & 21.5 / 39.8 / 56.2                            & 23.92$_{mp}$                \\
               & VMatcher-B                & 25.0 / 43.9 / 60.0                            & 21.5 / 39.5 / 56.1                            & 24.52$_{mp}$                \\
               & VMatcher-T                & 23.3 / 42.0 / 58.4                            & 19.7 / 37.5 / 54.1                            & \textbf{19.41}$_{mp}$       \\ \hline
               & ELoFTR$_{\text{Opt}}$     & 24.5 / 43.5 / 59.8                            & 20.0 / 38.2 / 54.7                            & 18.59$_{mp}$                \\
Semi-Dense-Opt & VMatcher-B$_{\text{Opt}}$ & \textbf{24.9} / \textbf{43.8} / \textbf{60.0} & \textbf{21.1} / \textbf{39.1} / \textbf{55.5} & 19.84$_{mp}$                \\
               & VMatcher-T$_{\text{Opt}}$ & 22.9 / 41.4 / 58.5                            & 19.5 / 37.1 / 53.7                            & \textbf{14.54}$_{mp}$       \\ \hline
\end{tabular}%
}
\caption{\textbf{Indoor Relative Pose Estimation Results on ScanNet \cite{Scannet}.} VMatcher variants are on par to other sparse and semi-dense methods, while VMatcher-T$_{\text{Opt}}$ offers the optimal balance between speed and accuracy.}
\label{tab:scannet}
\end{table}

\subsection{Visual Localization}\label{vl-sec}

\begin{figure*}[ht]
  \centering
  \includegraphics[width=\linewidth]{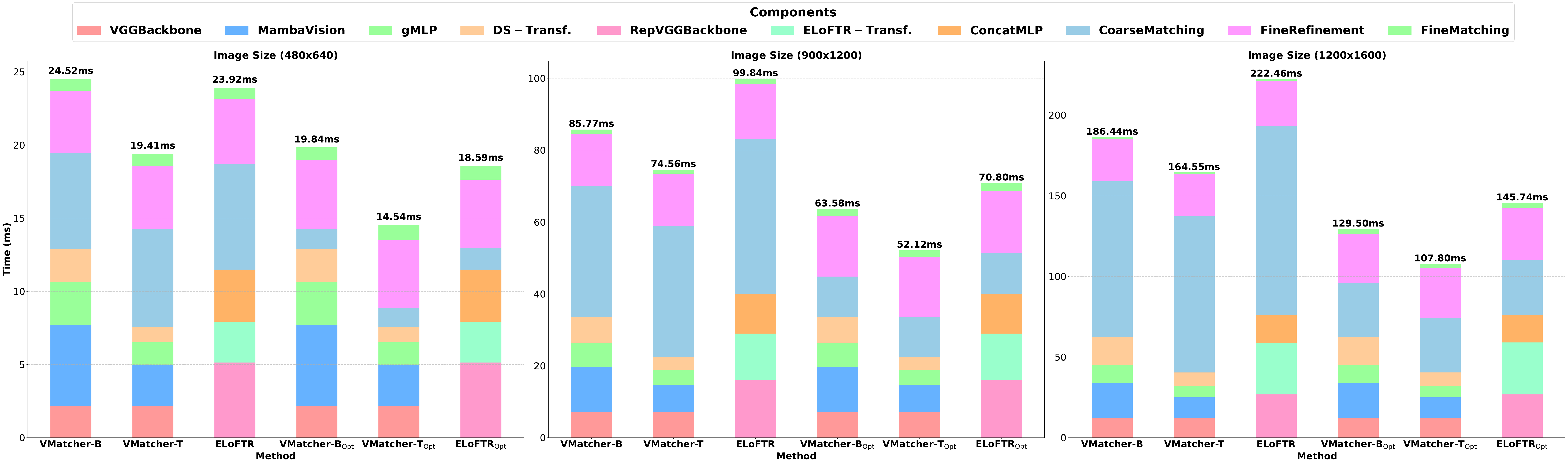}
  \caption{\textbf{Runtime Breakdown.} A runtime breakdown of VMatcher's pipeline across multiple image resolutions shows efficiency gains over ELoFTR \cite{Eloftr} as the resolution increases.}
  \label{fig:runtime}
\end{figure*}

Visual localization is evaluated on the Aachen v1.1 and InLoc datasets \cite{Aachen,inloc}. Aachen v1.1 \cite{Aachen} presents challenging viewpoint variations and illumination changes, including day-night transitions, while InLoc \cite{inloc} features indoor scenes with repetitive structures and textureless regions. Following \cite{LoFTR, Aspanformer, Eloftr, LightGlue}, the Hierarchical Localization (HLoc) framework \cite{Hloc} is employed for benchmarking. Results are reported for Aachen v1.1's day/night scenes and InLoc's DUC1/DUC2 test sets.

\noindent\textbf{Baselines.} Sparse methods include SuperPoint \cite{SuperPoint} with SuperGlue \cite{SuperGlue} and LightGlue \cite{LightGlue}, while for semi-dense methods VMatcher is evaluated against ELoFTR \cite{Eloftr}.

\noindent\textbf{Evaluation Protocol.} HLoc's \cite{Hloc} evaluation pipeline is followed for both sparse and semi-dense methods. For semi-dense methods, the longest edge of Aachen v1.1 and InLoc \cite{Aachen, inloc} images is resized to 1184 and 800 pixels, respectively. ELoFTR \cite{Eloftr} is evaluated under these conditions since no evaluation settings are reported. Results are reported as the percentage of poses within predefined distance and angular error thresholds. For Absolute Pose Estimation (APE) on the Aachen v1.1 dataset \cite{Aachen}, HLoc’s default configuration using PyCOLMAP \cite{colmap1, colmap2} is adopted. In contrast, due to the stochastic behaviour of PyCOLMAP on the InLoc dataset \cite{inloc}, affecting both sparse and semi-dense methods, APE is performed using PoseLib.

\noindent\textbf{Results.} From Tab. \cref{tab:aachen}, both sparse and semi-dense methods yield similar results on Aachen v1.1 \cite{Aachen}, with insignificant variations attributed to dataset saturation \cite{LightGlue}. Moreover, \cref{tab:inloc} shows that VMatcher performs on par with state-of-the-art feature matchers on InLoc \cite{inloc}, with differences being statistically insignificant due to limited query samples per test split \cite{LightGlue}. However, results highlight the computational advantage of VMatcher, particularly the optimised models achieving sparse-like runtime speeds.

\begin{table}[ht]
\centering
\resizebox{\columnwidth}{!}{%
\begin{tabular}{ccccc}
\hline
Matcher Type   & Method                    & \multicolumn{2}{c}{Aachen v1.1 Dataset}                                                       & Time (ms)                                                \\ \cline{3-4}
               &                           & Day                                           & Night                                         &                                                          \\ \cline{3-4}
               &                           & \multicolumn{2}{c}{(0.25m,$2^\circ$)/(0.5m,$5^\circ$)/(1.0m,$10^\circ$)}                      &                                                          \\ \hline
Sparse         & SP+SG                     & \textbf{90.3} / \textbf{96.2} / \textbf{99.4} & 76.4 / 90.1 / \textbf{100.0}                  & \multicolumn{1}{c}{209.50 (19.27/168.23$_{mp}$)}        \\
               & SP+LG                     & \textbf{90.3} / \textbf{96.2} / 99.2          & \textbf{77.5} / \textbf{91.6} / 99.5          & \multicolumn{1}{c}{\textbf{51.35} (19.27/32.08$_{mp}$)} \\ \hline
               & ELoFTR                    & 89.0 / 96.0 / 98.7                            & 77.0 / 91.1 / \textbf{99.5}          & \multicolumn{1}{c}{90.91$_{mp}$}                        \\
Semi-Dense     & VMatcher-B                & \textbf{89.4} / \textbf{96.5} / \textbf{99.0} & 77.0 / \textbf{91.6} / \textbf{99.5}          & \multicolumn{1}{c}{78.32$_{mp}$}                        \\
               & VMatcher-T                & 88.7 / 96.0 / 98.8                            & \textbf{77.5} / 91.1 / \textbf{99.5}                   & \multicolumn{1}{c}{\textbf{68.06}$_{mp}$}               \\ \hline
               & ELoFTR$_{\text{Opt}}$     & 89.0 / 95.5 / 98.4         & 76.4 / 91.1 / \textbf{99.0} & \multicolumn{1}{c}{60.13$_{mp}$}                        \\
Semi-Dense-Opt & VMatcher-B$_{\text{Opt}}$ & \textbf{89.3} / \textbf{96.1} / \textbf{98.8} & \textbf{77.5} / \textbf{91.6} / \textbf{99.0}          & \multicolumn{1}{c}{54.19$_{mp}$}                        \\
               & VMatcher-T$_{\text{Opt}}$ & 88.8 / 95.6 / 98.1                            & 77.0 / 91.1 / \textbf{99.0}                   & \multicolumn{1}{c}{\textbf{46.58}$_{mp}$}               \\ \hline
\end{tabular}%
}
\caption{\textbf{Aachen v1.1 Dataset \cite{Aachen} Visual Relocalization.}}
\label{tab:aachen}
\end{table}

\begin{table}[ht]
\centering
\resizebox{\columnwidth}{!}{%
\begin{tabular}{ccccc}
\hline
Matcher Type   & Method                    & \multicolumn{2}{c}{InLoc Dataset}                                                             & Time (ms)                           \\ \cline{3-4}
               &                           & DUC1                                          & DUC2                                          &                                     \\ \cline{3-4}
               &                           & \multicolumn{2}{c}{(0.25m,$2^\circ$)/(0.5m,$5^\circ$)/(1.0m,$10^\circ$)}                      &                                     \\ \hline
Sparse         & SP+SG                     & \textbf{46.0} / \textbf{68.7} / \textbf{79.8} & \textbf{57.3} / \textbf{77.9} / \textbf{79.4} & 153.12 (21.44/131.68$_{mp}$)        \\
               & SP+LG                     & 44.9 / 68.2 / 77.8                            & 56.5 / 73.3 / 77.9                            & \textbf{40.21} (21.44/18.77$_{mp}$) \\ \hline
               & ELoFTR                    & 52.5 / \textbf{74.2} / \textbf{86.4}          & \textbf{64.1} / \textbf{81.7} / \textbf{86.3} & 39.44$_{mp}$                        \\
Semi-Dense     & VMatcher-B                & \textbf{55.1} / \textbf{74.2} / 85.9                   & 61.1 / \textbf{81.7} / 85.5                   & 36.79$_{mp}$                        \\
               & VMatcher-T                & 53.5 / \textbf{74.2} / 85.4                   & 59.5 / 79.4 / 83.2                            & \textbf{32.52}$_{mp}$               \\ \hline
               & ELoFTR$_{\text{Opt}}$     & 50.5 / 72.2 / 84.3                            & 59.5 / 78.6 / 84.0                            & 32.75$_{mp}$                        \\
Semi-Dense-Opt & VMatcher-B$_{\text{Opt}}$ & 50.5 / 72.2 / 83.8                            & \textbf{61.1} / \textbf{79.4} / \textbf{84.7} & 30.09$_{mp}$                        \\
               & VMatcher-T$_{\text{Opt}}$ & \textbf{52.0} / \textbf{74.2} / \textbf{85.4} & 60.3 / 78.6 / 81.7                            & \textbf{24.82}$_{mp}$               \\ \hline
\end{tabular}%
}
\caption{\textbf{InLoc Dataset \cite{inloc} Visual Relocalization.}}
\label{tab:inloc}
\end{table}

\subsection{Runtime Breakdown}\label{Runtimebreakdown}

From \cref{hom-est-sec} to \cref{vl-sec}, it is evident that the runtime of VMatcher varies across experiments. Therefore, \cref{fig:runtime} presents a runtime breakdown on ScanNet \cite{Scannet} across multiple resolutions, providing a more detailed analysis. As a sequence-length-dependent model, the Mamba \cite{Mamba, Mamba2} architecture imposes a constant computational overhead of $\approx 0.8 ms$ per layer. At lower resolutions, VMatcher and its optimised variants achieve runtimes comparable to ELoFTR \cite{Eloftr}, with VMatcher-T being slightly more efficient. At higher resolutions, this overhead becomes negligible, allowing VMatcher to maintain competitive performance while improving efficiency across all variants.

\subsection{Ablation Study}\label{ablation}

\begin{table}[ht]
\centering
\resizebox{\columnwidth}{!}{%
\begin{tabular}{cccc}
\hline
Method & \multicolumn{2}{c}{MegaDepth Dataset} & Time (ms) \\ \cline{2-3}
& \multicolumn{2}{c}{AUC@$5^\circ$ / AUC@$10^\circ$  / AUC@$20^\circ$} & \\ \cline{2-3}
& LO-RANSAC & RANSAC & \\ \hline
VMatcher-B (Full Model) & 69.5 / 81.1 / 88.9 & 56.0 / 72.2 / 83.6 & 87.56$_{mp}$ \\
a) VGG Backbone only & 61.8 / 72.3 / 82.2 & 47.7 / 64.1 / 75.4 & 64.42$_{mp}$ \\
b) VGG + DS-Transformer (w/o MambaV + gMLP) & 64.9 / 75.6 / 84.1 & 49.9 / 65.9 / 77.6 &  69.51$_{mp}$\\
c) VGG + MambaVision + gMLP (w/o DS-Transformer) & 65.4 / 76.9 / 85.7 & 50.5 / 66.4 / 78.3 & 80.14$_{mp}$ \\
d) Full Model w/ Rep-VGG Backbone & 69.1 / 80.7 / 88.5 & 55.6 / 71.4 / 83.1 & 97.22$_{mp}$ \\
e) MLP replacing gMLP & 68.1 / 80.1 / 88.0 & 53.7 / 70.2 / 82.2 & 86.86$_{mp}$ \\
f) MLP after Attention & 67.7 / 79.9 / 87.8 & 53.4 / 70.0 / 82.0 & 89.77$_{mp}$ \\
g) Full Model w/o Self-Attention RoPE & 69.7 / 81.2 / 88.9 & 56.0 / 72.1 / 83.2 & 86.79$_{mp}$ \\ \hline
\end{tabular}%
}
\caption{\textbf{Ablation Study.} Analysing VMatcher's architecture.}
\label{tab:bitbybit}
\end{table}

\cref{tab:bitbybit} presents an ablation study of VMatcher, highlighting the impact of the hybrid Mamba-transformer architecture and key design choices. The VGG backbone alone (row a) shows significantly reduced performance, while adding either DS-Transformer or MambaVision+gMLP substantially improves results (rows b, c). However, neither component independently matches the full model's capabilities, confirming the effectiveness of the hybrid approach. Replacing the lightweight VGG backbone with RepVGG (row d) yields lower performance with higher computational overhead, validating the lightweight backbone choice in \cref{FeatureExtSec}. Substituting the Gated MLP with a standard Feed-Forward MLP (row e) offers minor runtime improvements but compromises accuracy. As detailed in \cref{dstransformersec}, adding an MLP (row f) post-attention negatively impacts both performance and efficiency. Notably, removing RoPE \cite{Rope} from self-attention (row g) slightly improves runtime and accuracy, which aligns with the hypothesis in \cref{dstransformersec} that Mamba's \cite{Mamba} inherent positional encoding makes additional positional encoding unnecessary.
\section{Conclusion}
\label{sec:conclusion}

This paper introduces VMatcher, a hybrid Mamba-Transformer semi-dense feature matcher that prioritises computational efficiency without sacrificing performance. Evaluations demonstrated that VMatcher achieves runtime gains over existing semi-dense matchers, with optimised variants reaching sparse-like processing speeds, marking a notable advancement in semi-dense feature matching efficiency. Future work may focus on further refining the architecture, aiming for greater computational efficiency while maintaining accuracy. This work hopes to pave the way for more efficient feature matching methodologies, contributing to the ongoing evolution of the field and offering a promising direction towards more practical and efficient solutions.
{
    \small
    \bibliographystyle{ieeenat_fullname}
    \bibliography{main}
}

\clearpage
% \setcounter{page}{1}
% \maketitlesupplementary
\section*{\Large{Appendix}}

\renewcommand{\thesection}{\Alph{section}}
\setcounter{section}{0}

\section{VMatcher Model Configurations}\label{vmatcherarchitecture}
VMatcher configurations introduced in \cref{sec:implementation_details} are designed to balance computational efficiency and matching accuracy by varying the model size:

\begin{itemize}
    \item \textbf{VMatcher-B} (Base):
    \begin{itemize}
        \item Architecture: 24 layers (9.5M parameters)
        \item Layer Pattern: [M G M$_{s}$ G S M G M$_{s}$ G C M G M$_{s}$ G S M G M$_{s}$ G C M G M$_{s}$ G]
    \end{itemize}
    
    \item \textbf{VMatcher-T} (Tiny):
    \begin{itemize}
        \item Architecture: 14 layers (6.9M parameters)
        \item Layer Pattern: [M G M$_{s}$ G S M G M$_{s}$ G C M G M$_{s}$ G]
    \end{itemize}
\end{itemize}

\noindent Where the architectural components are defined as follows:
\begin{itemize}
    \item M: MambaVision layer
    \item M$_{s}$: MambaVision$_{s}$ layer (MambaVision with transposed input and output is restored to its original orientation). Switches from row-major to column-major scanning.
    \item G: Gated Multi-Layer Perceptron (gMLP) layer
    \item S: Downsampled-Transformer Self-attention layer
    \item C: Downsampled-Transformer Cross-attention layer
\end{itemize}

% --------------------------------------------------

\section{Training Details}

All models were trained exclusively on the MegaDepth dataset \cite{Megadepth}, featuring 196 tourist landmarks with camera calibration, poses, and depth maps derived through COLMAP SfM and Multi-view Stereo (MVS) \cite{colmap1, colmap2}. Training used a learning rate of $1\times10^{-4}$ with the AdamW optimiser, a weight decay of 0.1, and a batch size of 1 over 30 epochs. Before attention layers, the feature vectors are interpolated (Bilinear) by a factor of 4. The loss weights \(\alpha\) and \(\beta\) in \cref{eq:total_loss} were set at 1.0 and 0.25, respectively. To accommodate more samples per batch, gradient accumulation was set to 8 for VMatcher-B and 32 for VMatcher-T. All training and evaluations were conducted on a single RTX 3090Ti GPU using PyTorch \cite{Pytorch}, with each epoch taking approximately 2 hours. 

% --------------------------------------------------

\section{Bidirectional Models}\label{bidirectionalmodels}
\begin{figure}[ht] 
    \centering
    \includegraphics[width=0.9\linewidth]{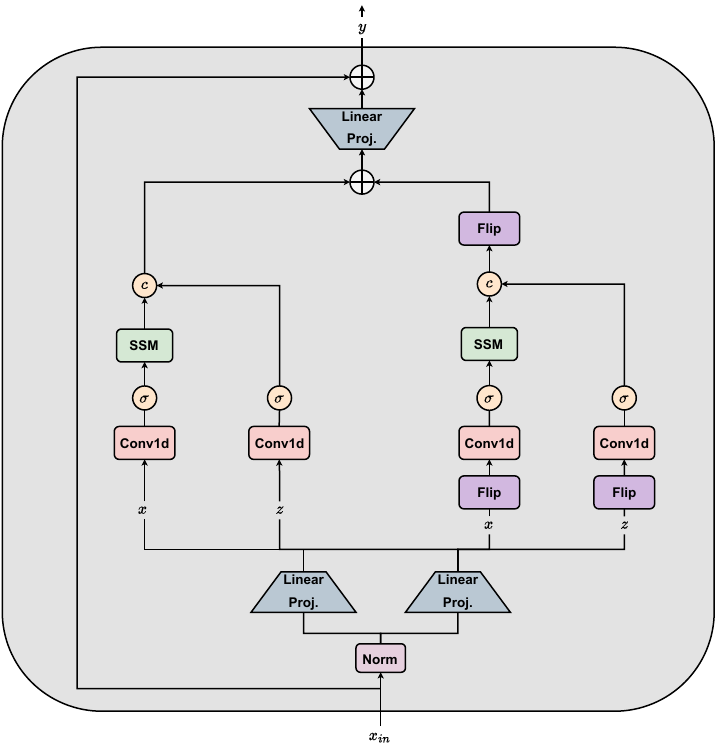}
    \caption{\textbf{Bidirectional MambaVision Block.}}  
    \label{fig:ssm_b}
\end{figure}

Two variants of VMatcher-B and VMatcher-T are introduced, where in the MambaVision layer \cite{MambaVision}, rather than scanning the feature vector in a Unidirectional manner (\emph{see} \cref{fig:VisionMambaBlock}), the scan is performed Bidirectionally as seen in \cref{fig:ssm_b}. For the Bidirectional models, the MambaVision layer computes the following:

\begin{equation}\label{eq.bi1}
    y_f = \text{MambaVision}(\text{Norm}(x_{in})).
\end{equation}

\begin{equation}\label{eq.bi2}
    y_b = \text{flip}(\text{MambaVision}(\text{flip}(\text{Norm}(x_{in})))).
\end{equation}

\noindent The outputs of Eqs. \ref{eq.bi1} and \ref{eq.bi2} are subsequently passed through the last linear layer, resulting in the final output of the MambaVision layer, as seen in Eq. \ref{eq.bi3}:

\begin{equation}\label{eq.bi3}
    y = x_{in} + \text{Linear}(y_f + y_b).
\end{equation}

As seen in \cref{tab:bitab}, no substantial improvement is observed between the unidirectional and bidirectional models, aside from the increase in inference time. This aligns with the findings in other vision tasks \cite{MambaVision, vim}, where bidirectional scans offer limited practical benefit over unidirectional ones.

\begin{table}[ht]
\centering
\resizebox{\columnwidth}{!}{%
\begin{tabular}{ccccc}
\hline
Method          & \multicolumn{3}{c}{MegaDepth Dataset}           & Time (ms) \\ \cline{2-4}
                & AUC@$5^\circ$ & AUC@$10^\circ$ & AUC@$20^\circ$ &           \\ \hline

VMatcher-B        & 69.6 &  81.1 & 88.9 & 87.56$_{mp}$   \\

VMatcher-B-Bi & 69.3 &  81.0 & 88.8 & 97.67$_{mp}$   \\ \hline

VMatcher-T  & 69.4 & 81.0 & 88.8 & 79.46$_{mp}$   \\

VMatcher-T-Bi & 69.4  & 80.6 & 88.6  & 86.25$_{mp}$ \\ \hline   

\end{tabular}%
}
\caption{\textbf{Unidirectional and Bidirectional VMatcher model comparison.}}
\label{tab:bitab}
\end{table}

\begin{figure*}[ht]
  \centering
  \includegraphics[width=\linewidth]{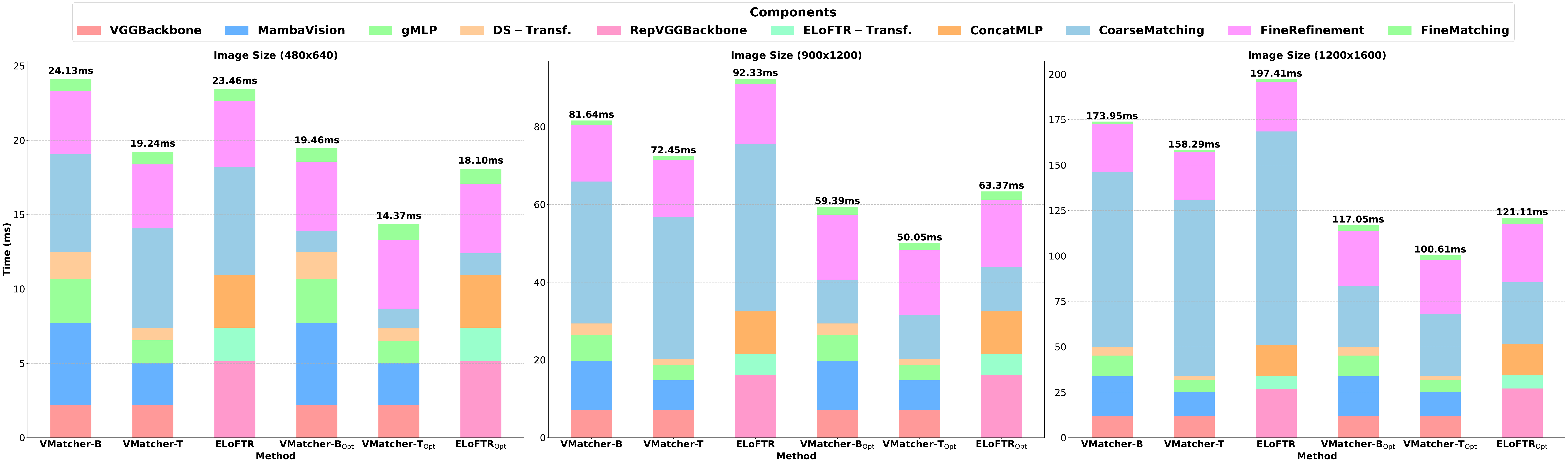}
  \caption{\textbf{Runtime Breakdown Extended.} Flash Attention \cite{FlashAtt} enabled during inference time measurement on ScanNet \cite{Scannet}.}
  \label{fig:runtimeextend}
\end{figure*}

% --------------------------------------------------
\vspace{-0.8cm}
\section{Design Choices}\label{DesignChoices}

\subsection{Downsampled Transformer}\label{DTsup}

Preserving information during downsampling is crucial, prompting experiments with various methods, including Depth-wise Convolution, Area Interpolation, Max Pooling, and Average Pooling. While Bilinear Interpolation is less efficient for downsampling compared to other methods, as seen in \cref{tab:downsamplingtimes}, experiments revealed that Max or Average pooling resulted in the most degradation in performance by $\approx 1.5-2\%$. Depth-Wise convolution was avoided despite its efficiency to prevent introducing additional learnable parameters, as the downsampling quality would depend on the learnt convolutional kernels. Area Interpolation performed comparably to Bilinear Interpolation but showed a slight performance drop compared to Bilinear $\approx 0.5-1.0\%$. Given that the difference in runtime is negligible in the context of the complete pipeline, Bilinear Interpolation was ultimately chosen for its superior performance-efficiency trade-off.

\begin{table}[ht]
  \centering
  \resizebox{0.6\columnwidth}{!}{%
  \begin{tabular}{cc}
  \hline
  Downsampling Method    & Time (ms) \\ \hline
  Bilinear Interpolation & 0.072     \\
  Area Interpolation     & 0.042     \\
  Depth-wise convolution & 0.035     \\
  Max pooling            & 0.033     \\
  Average pooling        & 0.036     \\ \hline
  \end{tabular}%
  }
  \caption{\textbf{Downsampling methods runtimes.}}
  \label{tab:downsamplingtimes}
\end{table}

\begin{figure*}[ht]
    \centering
    \includegraphics[width=\linewidth]{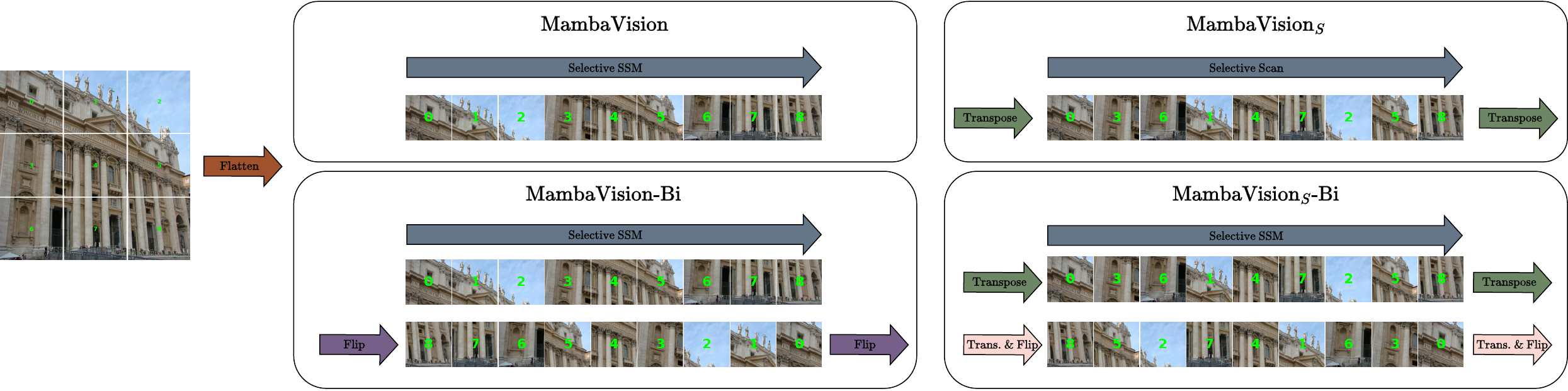}
    \caption{\textbf{VMatcher MambaVision layers scan directions visualisation.}}
    \label{fig:scan_vis}
\end{figure*}

\subsection{Mamba Vision}\label{mvsup}

While numerous Mamba \cite{Mamba} variants have been proposed for Computer Vision tasks; however, existing methods either prioritise high accuracy or fast inference speed, often at the cost of performance.  
Initially, the architecture was used Vanilla Mamba \cite{Mamba} in both bidirectional and unidirectional configurations. However, Vanilla Mamba proved less efficient than Mamba Vision \cite{MambaVision}, with nearly $1.5 \times$ the runtime and slightly lower performance. Given that the goal is to optimise both runtime and performance, MambaVision \cite{MambaVision} was chosen as the variant for VMatcher's architecture.

% --------------------------------------------------

\section{Rotation Invariance}

\begin{figure}[H]
    \centering
    \includegraphics[width=0.8\linewidth]{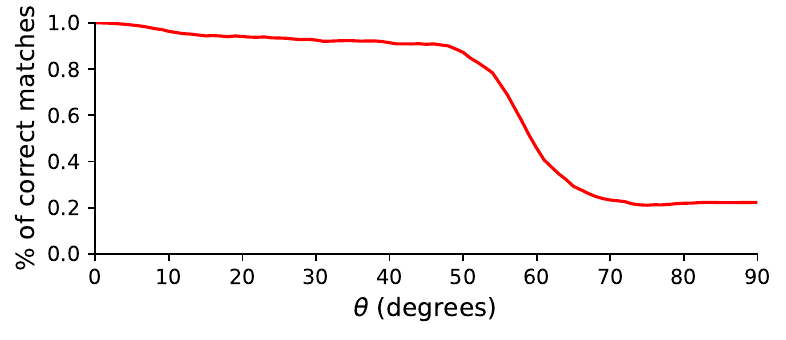}
    \caption{\textbf{Rotation Invariance evaluation on the IMC 2020 \cite{IMC2020} validation set.}}  
    \label{fig:rot}
\end{figure}

\noindent DISK's \cite{DISK} rotation invariance evaluation was performed on the Image Matching Challenge (IMC) 2020 \cite{IMC2020} validation set. For each angle $\theta$, 36 images are randomly selected and matched with their rotated counterparts. The ratio of correct matches, defined as those with a reprojection error below 3 pixels, is then computed. Results are shown in \cref{fig:rot}.

% --------------------------------------------------

\section{Runtime Breakdown extended}

VMatcher does not benefit from the usage of Flash Attention \cite{FlashAtt} compared to \cite{LoFTR, Eloftr, Aspanformer, LightGlue, SuperGlue}, due to the limited number of Transformer layers \cite{transformer} in its architecture. However, for fairness, \cref{fig:runtime} was extended by enabling Flash Attention during inference time measurement. As seen in \cref{fig:runtimeextend}, VMatcher maintains its superior inference speed, even with Flash-Attention \cite{FlashAtt} enabled.

% --------------------------------------------------

\section{VMatcher Scans Visualization}

\cref{sec:implementation_details,vmatcherarchitecture}  described VMatcher's architecture, which incorporates MambaVision and MambaVision$_{s}$ layers. Fig. \ref{fig:scan_vis} visualises their distinct scanning patterns using a simple $3\times3$ image example. The MambaVision layer processes the flattened image in row-major order, while MambaVision$_{s}$ first transposes the image to enable column-major scanning before restoring its original orientation. For bidirectional models introduced in Sec. \ref{bidirectionalmodels}, MambaVision-Bi scans both the flattened image and its flipped copy, then returns the flipped version to its original orientation as shown in Fig. \ref{fig:ssm_b}. Similarly, MambaVision$_{s}$-Bi transposes the image, scans both the transposed image and its flipped version, then restores both to their original orientation.

% --------------------------------------------------

\section{Limitations}

\begin{itemize} 
    \item The VMatcher model utilises the Mamba architecture, where a portion of the Mamba model is implemented in Triton \cite{Triton}, requiring a Graphics Processing Unit (GPU). While powerful hardware is common in state-of-the-art feature matching methods, this reliance on GPUs and modern hardware is believed to remain a general limitation. 
        
    \item While VMatcher offers significant runtime improvements, the efficiency of the Mamba \cite{Mamba} architecture depends on the sequence length. For smaller sequences, its runtime is comparable to Transformer-based models, but for larger sequences, it demonstrates marked efficiency gains. As Computer Vision tasks increasingly involve higher-resolution images, this advantage is expected to become more prominent. Moreover, as a relatively new architecture, Mamba \cite{Mamba} is likely to benefit from future optimisations, following a trajectory similar to the steady improvements seen in Transformer-based models.
\end{itemize}

% --------------------------------------------------

\section{Matching Examples}
\begin{figure*}[ht]
    \centering
    \includegraphics[width=\textwidth]{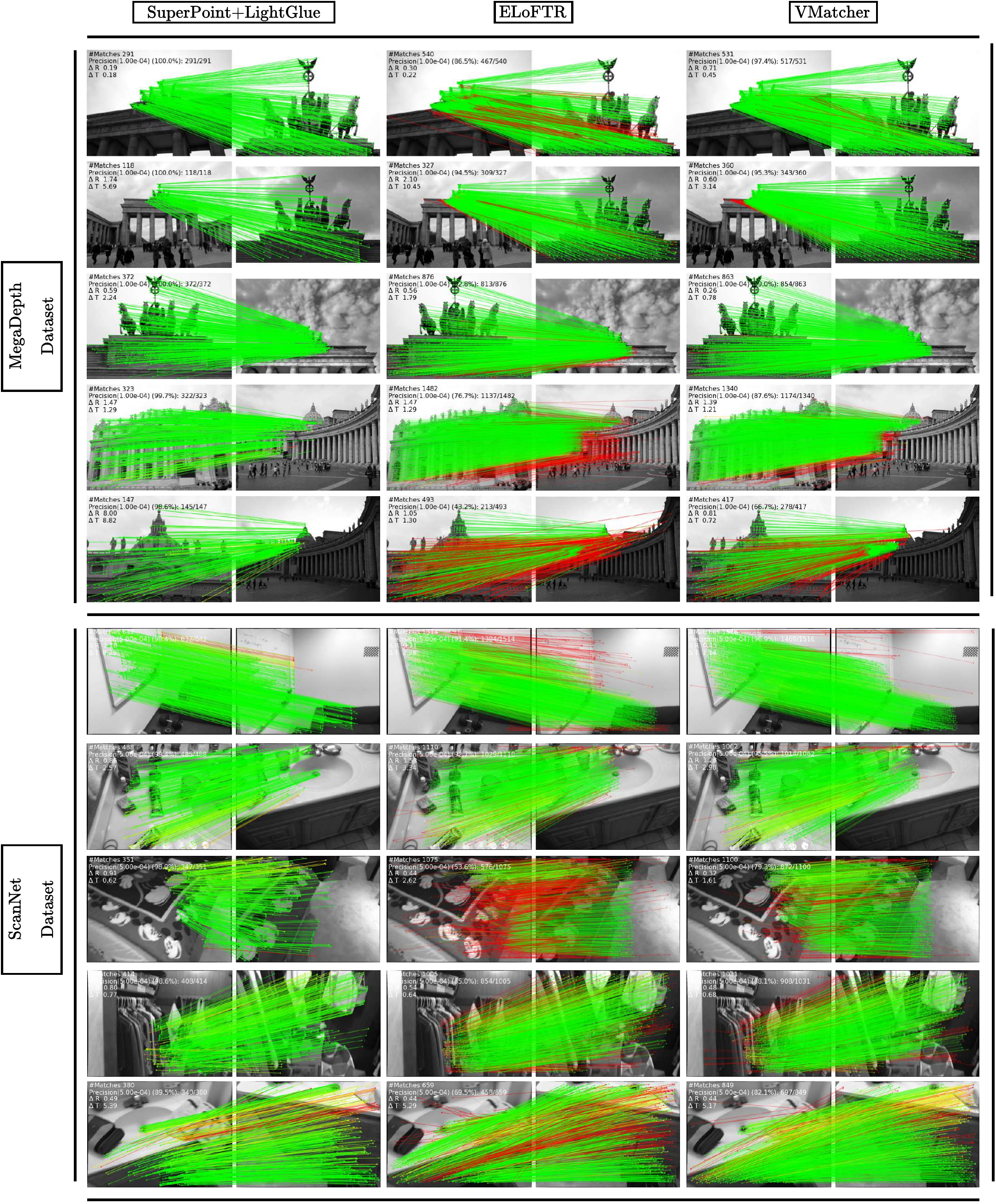}
    \caption{\textbf{Matching Examples on MegaDepth and ScanNet \cite{Megadepth,Scannet} datasets.}}  
    \label{fig:visual examples}
\end{figure*}

\cref{fig:visual examples} provides visual examples on the MegaDepth and ScanNet datasets \cite{Megadepth, Scannet}. Mismatches are not filtered out in the figures to present an unbiased demonstration.

\end{document}